The Metacognitive Bottleneck: Japanese Riddles Reveal Fundamental Limits of Machine Insight and Self-Evaluation in Reasoning AI

Masaharu Mizumoto,[1] Dat Nguyen,[2] Zhiheng Han,[1] Xingfu Li,[1] Yo Nakawake,[1] Le Minh Nguyen[2]

1 School of Knowledge Science, Japan Advanced Institute of Science and Technology

2 School of Information Science, Japan Advanced Institute of Science and Technology

**Abstract:** Benchmark saturation and training-data contamination increasingly obscure whether reported gains in large language models (LLMs) reflect genuine advances in reasoning or familiarity with recurring patterns in benchmark problems. We introduce the NazoNazo Benchmark, a renewable and extensible evaluation dataset derived from Japanese children's riddles that isolates a specific class of reasoning processes: insight-like representational restructuring and metacognitive evaluation. Rather than modeling reasoning in general, these tasks provide a focused test of failure modes that are difficult to detect in standard benchmarks.

We curate 201 riddles and establish a human reference on a 120-item subset (n = 126; mean accuracy 52.9%). The benchmark is fully open, low-cost to refresh, and designed for continual evaluation under reduced contamination risk. We evaluate 38 frontier LLMs (2023–2025) under a strict retrieval-free, zero-shot protocol. On the human-comparison subset, non-reasoning models achieve 7.6% accuracy and reasoning-oriented models reach 17.6%, compared with a human mean of 52.9%, although performance varies substantially across models.

Beyond accuracy, qualitative analysis of model-generated thought-logs identifies a distinctive failure mode, which we call *verification failure*: models generate a correct intermediate candidate but fail to endorse it as their final answer. This dissociation between candidate generation and endorsement reveals a metacognitive bottleneck: across the models with usable thought-logs, verification failures account for between 5% and 39% of a model's incorrect answers. By isolating the gap between generation and verification, this work provides a practical framework for diagnosing reasoning reliability and suggests concrete directions for improvement, including better calibration, structured verification, and stopping mechanisms.

## 1. Introduction

Benchmark saturation, rapid model turnover, and increasing overlap between training corpora and public test sets have made it progressively harder to interpret reported gains in large language model (LLM) "reasoning" (Maslej et al., 2025; Reuel et al., 2024). Strong improvements on widely used evaluations—often aided by inference-time procedures such as self-consistency or other test-time search—can reflect familiarity with formats or inadvertent exposure to items rather than transferable generalization (Guo et al., 2025; Latif et al., 2024; Wang et al., 2022). Compounding this problem, single-number benchmark scores frequently conflate distinct functions that matter for reliability in real use: producing a plausible candidate answer versus recognizing, validating, and deciding to stop on the correct one. This creates a measurement challenge at the intersection of AI evaluation and psychological measurement.

A particularly revealing class of tasks is insight-like problems, where progress depends on representational change rather than incremental computation (Knoblich et al., 1999; Ohlsson, 1992; Wiley & Danek, 2024). In humans, insight is commonly associated with impasse and sudden restructuring processes such as constraint relaxation and chunk decomposition, and performance is shaped by monitoring and control processes as much as by knowledge (Knoblich et al., 1999; Ohlsson, 1992; Wiley & Danek, 2024). Accordingly, errors can also arise from failures of verification, stopping, or confidence calibration—cases where a solver may generate a correct (or near-correct) candidate yet fail to endorse it (Gangemi et al., 2015; Nadurak, 2023; Thompson et al., 2011), though such cases are relatively rare in humans. This separation between candidate generation and candidate evaluation suggests an analogous vulnerability for LLMs: even when a system can sometimes construct the correct solution, it may not reliably select, verify, and commit to that solution as its final response. We address this gap with traditional Japanese children's riddles (*nazonazo*), a compact family of short (typically single-sentence) wordplay riddles that usually induce impasse and require representational shift.

Thus, nazonazo riddles, if not representative of all forms of reasoning, provide a specific but important class of reasoning processes: insight-like restructuring, in which success depends on representational change and metacognitive control rather than on incremental computation or knowledge retrieval. Many real-world reasoning failures, including creative problem solving, debugging, and hypothesis revision in scientific discovery, involve similar restructuring dynamics.

Nazonazo riddles are widely known among the Japanese, who are familiar with them from childhood (see SI Appendix 1). Just as other insight-like problems, nazonazo items typically require solvers to engage in trial-and-error exploration, relaxing unnecessary self-imposed constraints and transforming the original framing, or decomposing perceptual "chunks" into parts and recombining them into a new representation, often culminating in a sudden "aha!" moment (see Fig. 1 for an illustrative example).

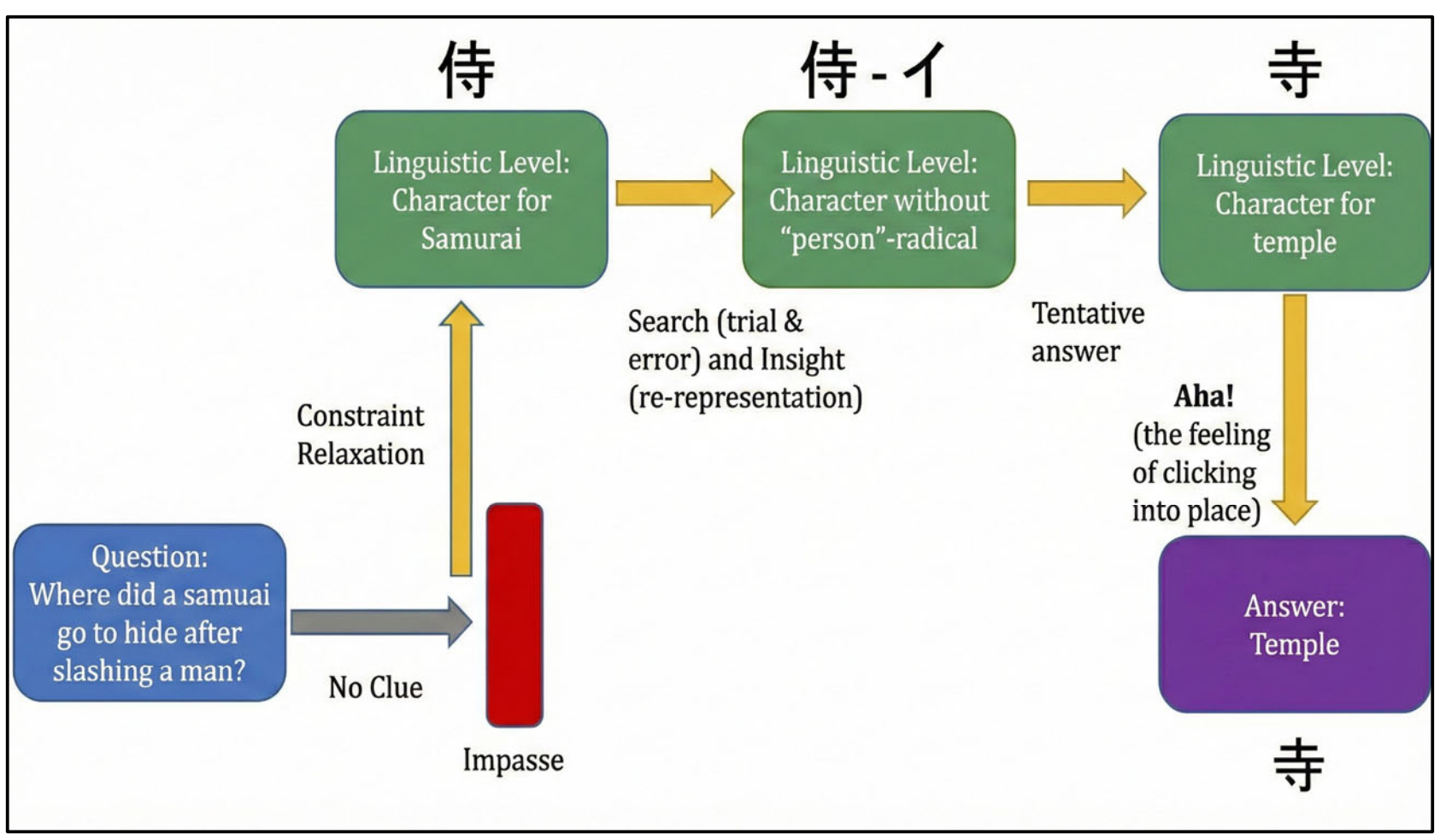

**Fig. 1. Illustration of insight problem-solving structure.** (Example decomposition/constraint-relaxation pathway yielding a sudden solution experience.)

Nazonazo items typically fall into four types: Dajare (puns), Goro awase (number rephrase), Kanji (character decomposition), and Hidden instruction, or mixtures thereof (see Table S1 in SI Appendix for details). Because such items are low-cost to refresh and easy to rotate, they offer a practical path to continual measurement with reduced contamination risk, and their renewability may also help preserve *discriminative headroom*, the remaining distance from saturation, where accuracy is near the ceiling, between-model variance becomes very small, and successive models show little or no improvement on newly created items, which is distinct from the direct measure of contamination.

Nazonazo complements existing puzzle and riddle evaluations, including rebus norms (Threadgold et al., 2018), symbol understanding tests (Gritsevskiy et al., 2024), English riddle datasets (Lin et al., 2021), joke-completion insight tasks (Bower, 2020), and multimodal puzzle benchmarks (Lee et al., 2025), by providing a non-English, continually refreshable instrument designed for repeated use over time. This class of tasks is especially diagnostic because it minimizes confounds from knowledge retrieval and focuses on search, restructuring, and evaluation processes that are otherwise difficult to isolate in standard benchmarks. Unlike many existing benchmarks, however, nazonazo tasks are also independent of reliance on memorized English-language patterns.

Here we curate the NazoNazo Benchmark, consisting of 201 nazonazo riddles from a public source (see SI Appendix 1) and define a 120-item human-comparison subset. We evaluate 38 leading LLMs released between 2023 and 2025 under a strict retrieval-free, zero-shot protocol intended to isolate internal problem-solving from external lookup. A human reference study provides an empirical upper reference point on the subset, while model accuracy remains substantially below the human reference, including those of many systems marketed as reasoning-oriented, although the strongest model approaches the observed human range.. Beyond accuracy, we analyze model-produced reasoning traces as qualitative diagnostics in the spirit of protocol analysis (Ericsson & Simon, 1993), while not assuming such traces are fully faithful or uniformly informative (Arcuschin et al., 2025; Chen et al., 2025; Deng et al., 2025).

In qualitative examination, across models we identified repeated instances of a "last-mile" limitation: verification failure, which we later quantify (Table 2). Models may generate the correct candidate during intermediate reasoning yet fail to endorse it, continuing exploration and ultimately responding incorrectly, which poses a challenge to the common claim that chain-of-thought traces are merely post hoc reconstructions. This dissociation between generative reasoning and metacognitive-style endorsement helps reconcile a common paradox in contemporary LLM use, such as occasional striking solutions alongside unreliable self-checking.

Here, we use "metacognitive" in a functional and behavioral sense, referring to monitoring, evaluation, and stopping dynamics rather than to subjective awareness. Framed in this way, verification and stopping emerge as first-class evaluation targets. Together, these results provide (i) a continually refreshable instrument for tracking progress in insight-like reasoning under reduced contamination risk, (ii) a behavioral substrate for comparing human and machine signatures of restructuring, monitoring, and stopping under matched conditions, and (iii) a concrete mechanism by which prompted self-verification can fail without external signals or structured checking procedures.[1]

## Results

**Human reference performance.** Out of 201 nazonazo riddles, we constructed a 120-item subset for direct human comparison. A pilot survey collected human responses to the 120-item subset (n=126), yielding a mean accuracy of 52.9% (95% CI: 46.6–59.2%) and a Gaussian item-level distribution (Fig. 2.1) and a bimodal score distribution (Fig. 2.2). The observed bimodality is consistent with prior accounts of performance patterns in insight problem solving and Aha!-related restructuring dynamics (Metcalfe, 1986; Metcalfe & Wiebe, 1987; Wiley & Danek, 2024). These results provide a human reference distribution against which model performance on the matched subset can be interpreted. Gold answer variants were established during the pilot to enable automated scoring in the model study. (For the details of this pilot study, see SI Appendix 2 and Table S2 there.)

[1] An earlier version of this paper can be found at: https://arxiv.org/abs/2509.14704. All relevant materials can be found at: DOI: 10.5281/zenodo.17019050.

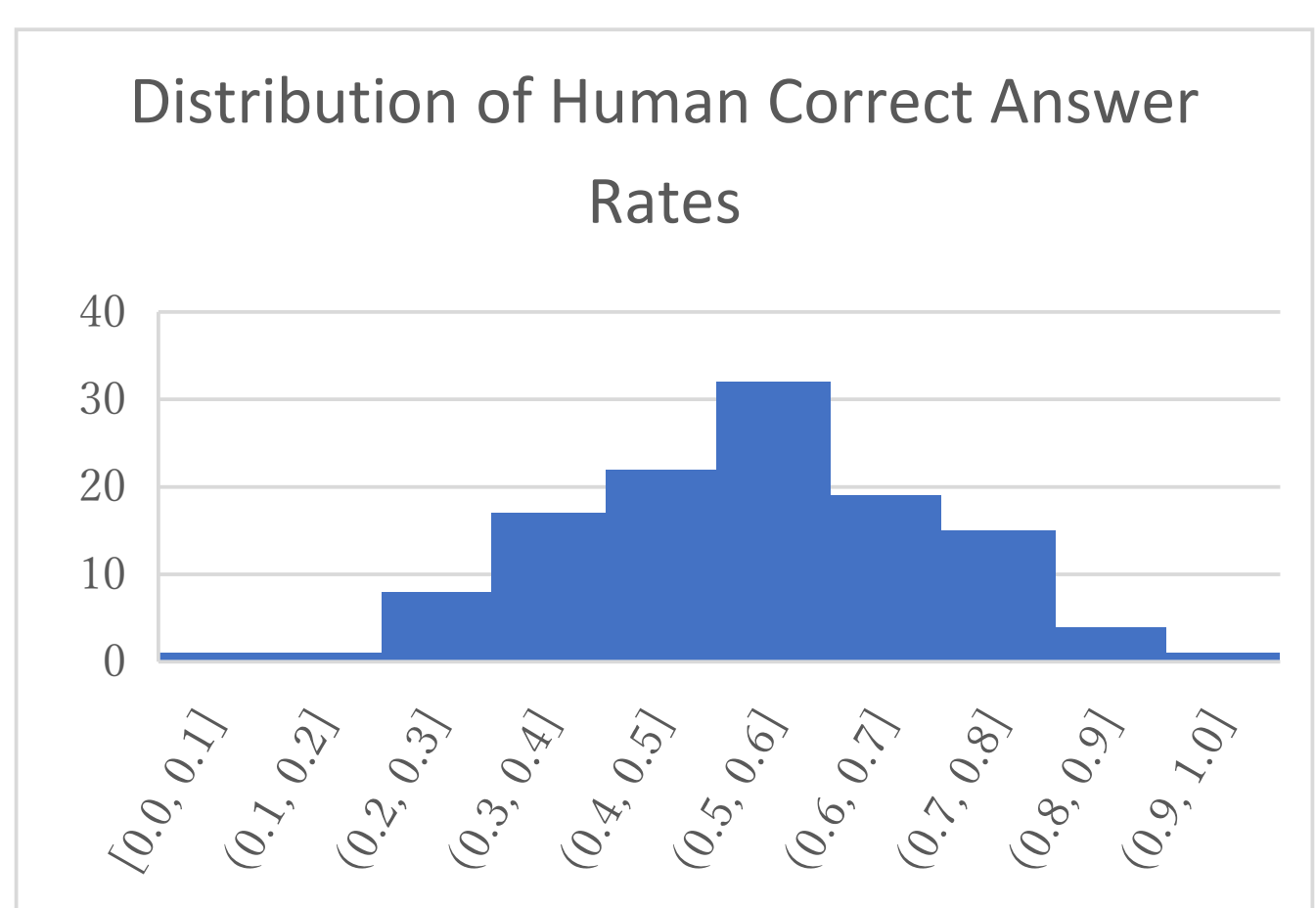

**Fig. 2.1: [2]Distributions of Item-Level Difficulty for Humans.** (At the item level, the histogram of human accuracies across the 120 riddles shows that the set was balanced between relatively easy and relatively difficult questions.)

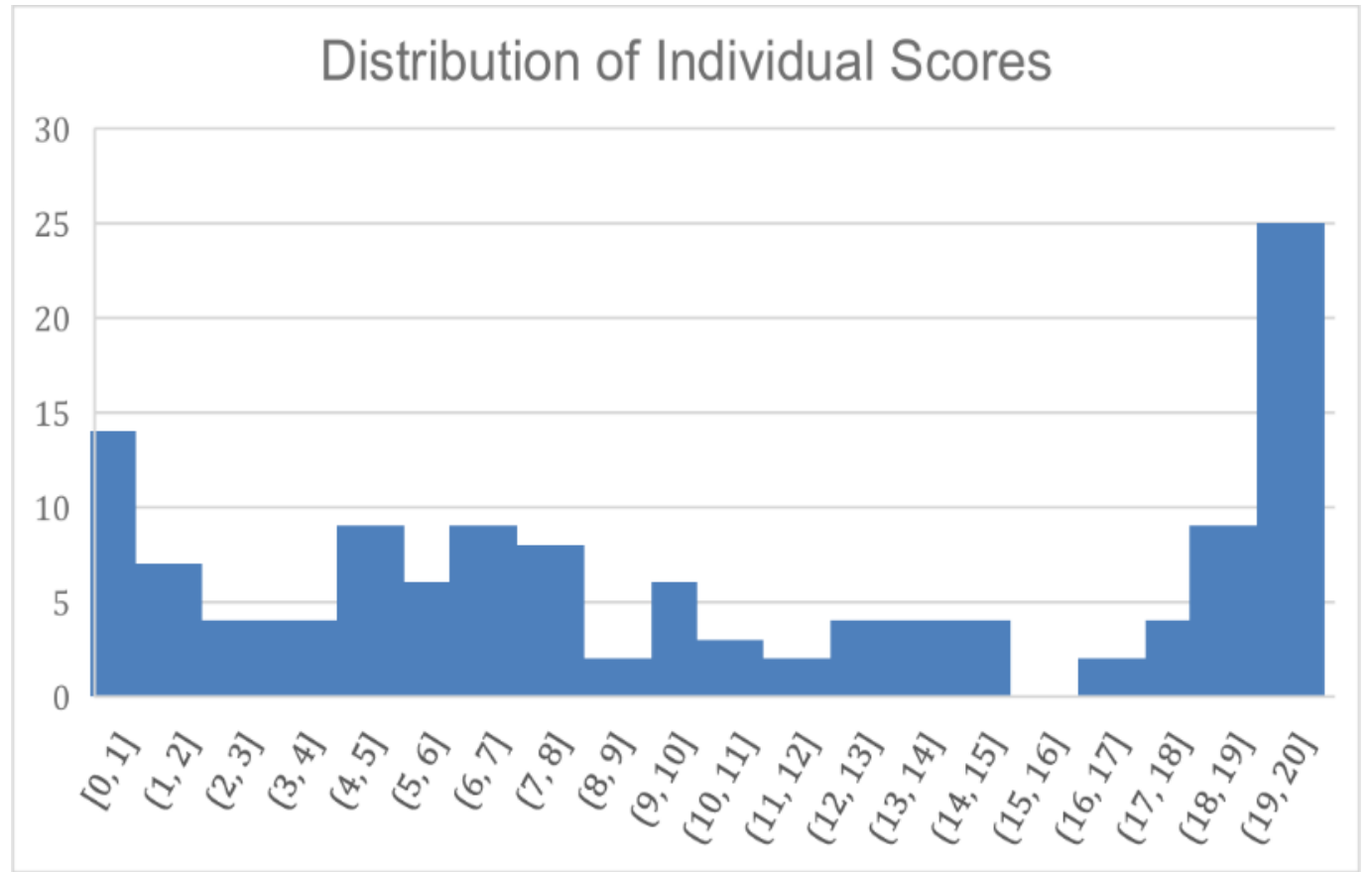

**Fig. 2.2 Human score distribution on the 120-item subset. (**In a pilot survey (n=126), participants answered 20 riddles sampled from items 1–120; total scores (out of 20) show clear bimodality with peaks at 0 and 20.)

**Most frontier LLM performance remains far below humans without retrieval.**

We evaluated 38 frontier LLMs using a strict, retrieval-free, zero-shot protocol on the 120-item subset and then on the full 201 items. On the human-comparison subset, non-reasoning models averaged 7.6% accuracy while reasoning models averaged 17.6% (Fig. 3), compared with a human mean of

[2] The study was approved by the Knowledge Science Ethics Council at Japan Advanced Institute of Science and Technology (Approval No. KSEC-G2025022103). All participants provided informed consent before participation.

52.9% (95% CI: 46.6%–59.2%). Thus, most evaluated models remained substantially below the human reference.

GPT-5 was a notable exception. It was the highest-performing and most recently released model in the evaluation, and its observed accuracy fell within the 95% confidence interval of the human mean. This descriptive overlap should not be interpreted as evidence of statistical equivalence, because we did not conduct formal model-by-model comparisons between individual LLMs and the human sample. GPT-5 also appeared as a high-performing outlier relative to the other reasoning-oriented models.

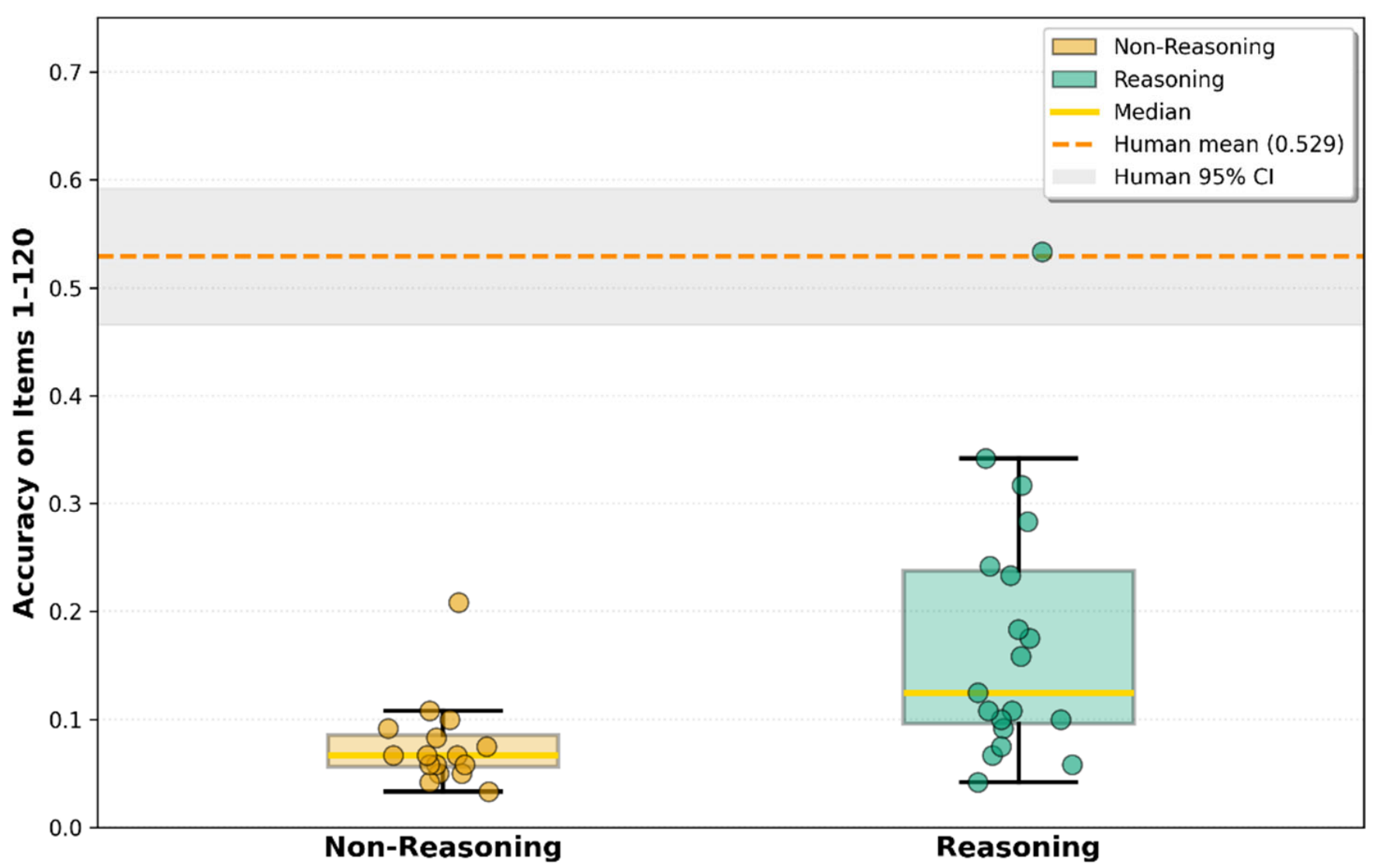

**Fig. 3. Model performance on the 120-item subset with human reference. Each point represents the accuracy of one model.** Dashed line indicates human mean; shaded region indicates the human 95% CI. GPT-5 is the high-performing point within the human confidence interval.

On the full 201 items, pooled item-level analysis estimated an odds ratio of 2.29 (95% CI: 1.99–2.64; $p < 0.0001$) for reasoning versus non-reasoning models, with population-level accuracies of 9.42% and 19.25%, respectively (Fig. 4).

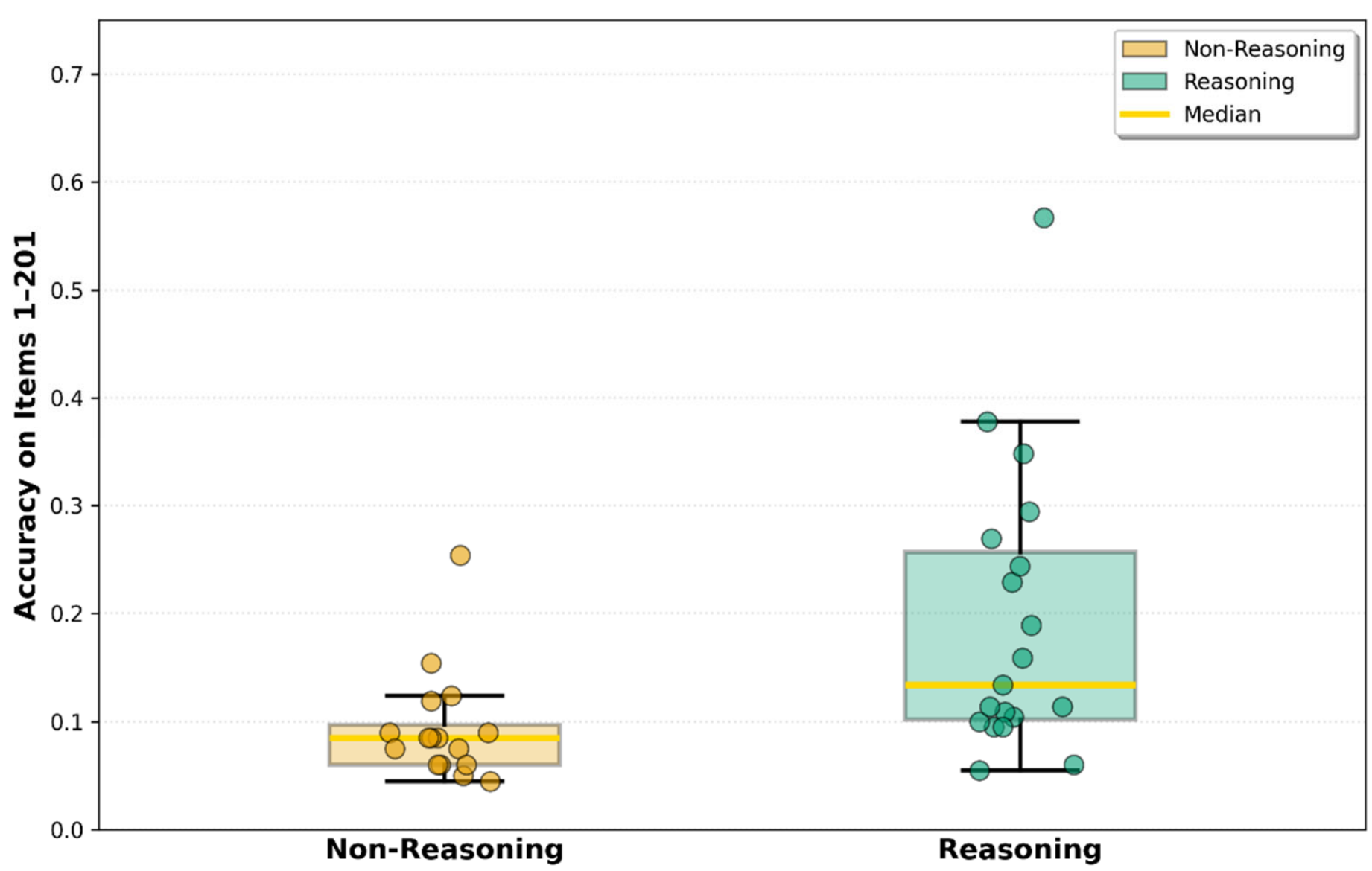

**Fig. 4. Non-reasoning vs. reasoning model performance on all 201 items.** Each point denotes a model's accuracy; boxes summarize distributions, with the median shown as a line.

Across models, parameter count did not predict accuracy, whereas release date correlated positively, consistent with the view that system design and training procedures can matter more than scale alone (Maslej et al., 2025). Model accuracies are heavily skewed toward low values (Fig. 5).

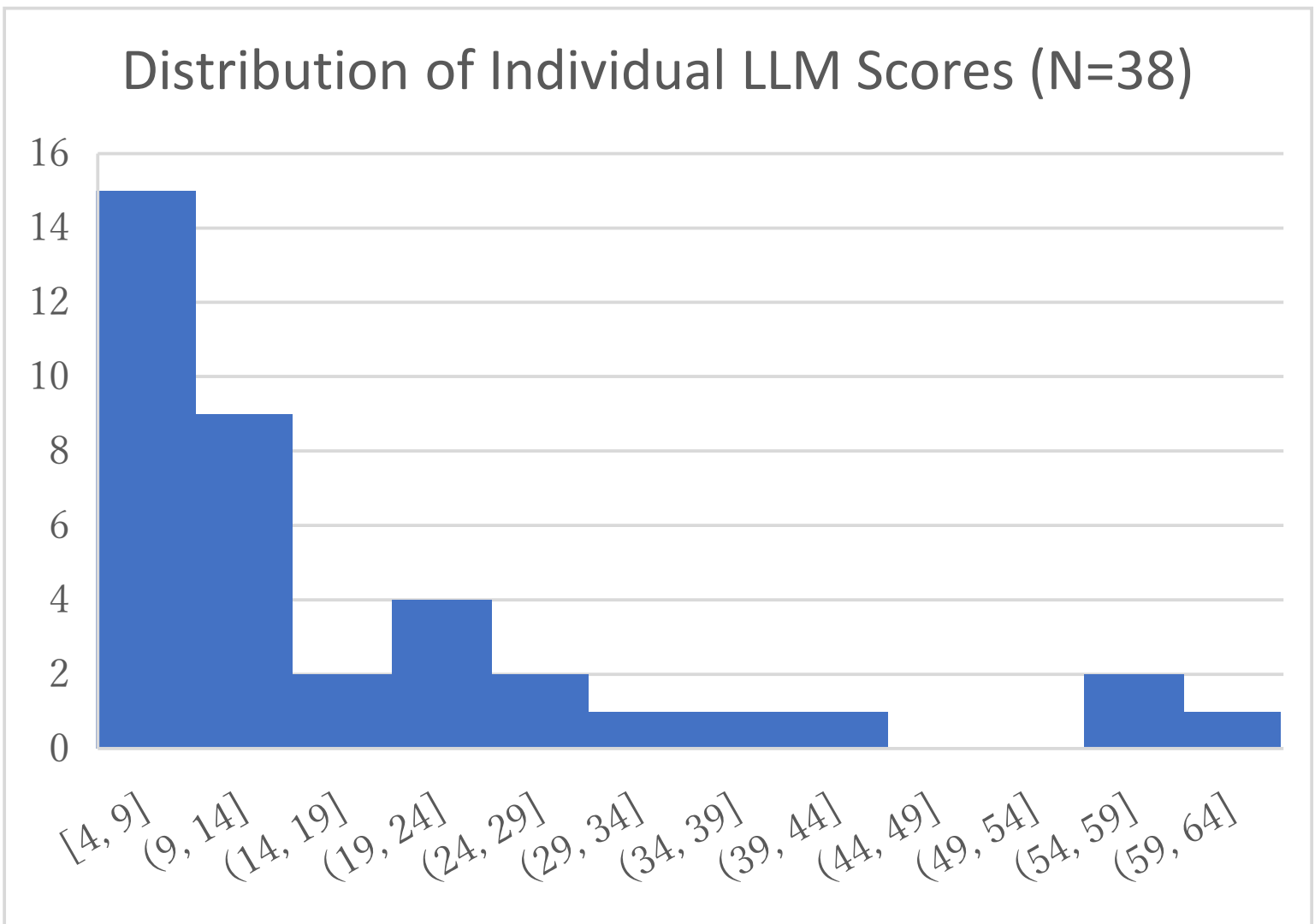

**Fig. 5. Distribution of individual LLM performance on the 120-item subset.**
Histogram of model accuracies across all 38 evaluated models (items 1–120).

Most individual items remain extremely difficult for LLMs (Fig. 6), in contrast with the Gaussian distribution for humans (reported in SI Appendix).

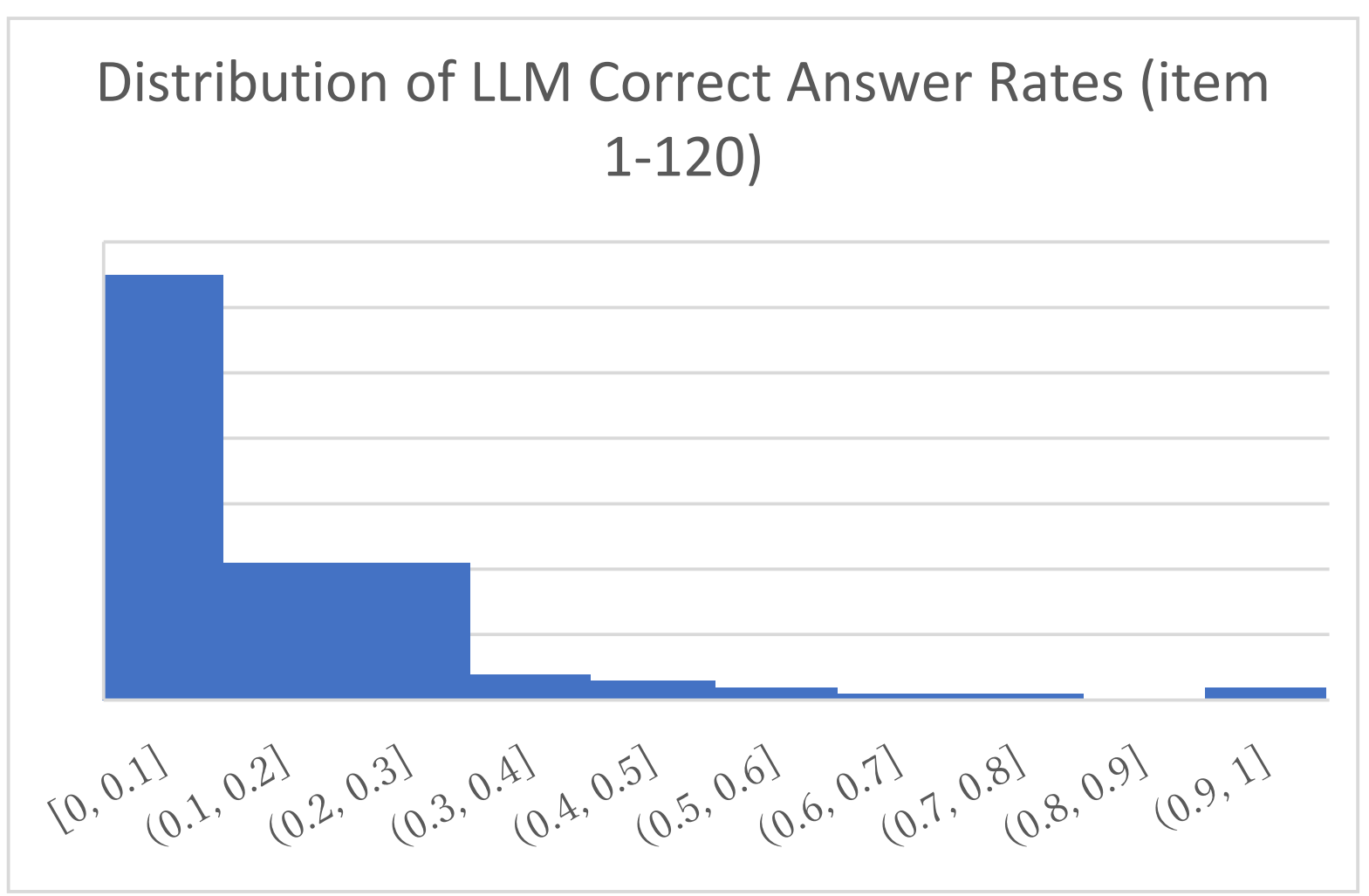

**Fig. 6. Item difficulty distribution for LLMs on the 120-item subset.** Histogram of per-item correctness rates across models (items 1–120).

**Thought-log patterns: language choice and verification failures.**

We qualitatively analyzed thought logs from representative reasoning models as diagnostic traces of failure mechanisms. Across models, two recurring behaviors stand out: (i) the language used during intermediate reasoning (English vs. Japanese) and (ii) breakdowns in solution endorsement and stopping, even when a correct candidate is generated during search.

First, the language choice in thought logs did not reliably improve accuracy. Reasoning in English rather than Japanese (or vice versa) yielded inconsistent effects across models and items, suggesting that the dominant bottleneck is not surface-level comprehension alone but the ability to restructure the problem representation and to evaluate intermediate hypotheses (Knoblich et al., 1999; Thompson et al., 2011). Table 1 summarizes language choice in thought logs and accuracy by language for representative reasoning models.

| Model (overall) | Language proportions | Lines per question | Accuracy by language | Test |
|---|---|---|---|---|
| DeepSeek-R1-0528 (22.9%) | Japanese: 100% (199/199); 2 missing | Japanese: 10,397 (52.3/q) | Japanese: 23.1% (46/199) | - |
| Gemini-2.5-flash-preview (10.9%) | Japanese: 6.0% (12); English: 94% (188); both used in 1 item | Japanese: 2,511 (209.3/q); English: 2,791 (14.8/q) | Japanese: 0% (0/12); English: 11.7% (22/188) | p = 0.369 |
| Claude-opus-4 (30.3%) | Japanese: 21.9% (44); English: 78.1% (157) | Japanese: 2,021 (45.9/q); English: 7,139 (45.5/q) | Japanese: 36.7% (16/44); English: 28.7% (45/157) | p = 0.356 |
| Grok-4 (37.6%) | Japanese: 57.2% (111); English: 42.8% (82); 7 missing; 1 with no log | Japanese: 2,691 (24.2/q); English: 2,443 (29.8/q) | Japanese: 29.5% (33/112); English: 48.8% (40/82) | p = 0.0071 |

**Table 1. Summary of model performances by thought-log language.** Language proportions indicate the share of thought logs in Japanese vs English; accuracy is reported separately by language. Two-tailed Fisher's exact tests compare correctness rates by language within each model when both languages occur. Missing thought logs are excluded from denominators where applicable.

Second, we identified multiple explicit cases of a candidate–endorsement dissociation that we term *verification failure*: the model explicitly proposes a correct candidate during search but does not commit to it as the final answer, continues exploring, and ultimately responds incorrectly. (These cases occurred in more than one model and on multiple benchmark items; representative cases and item identifiers are reported in SI Appendix 3.) Table 2 quantifies this pattern across the four models with usable thought-logs: verification failures accounted for between 5.23% and 39.34% of each model's incorrect answers, highest for Grok 4 and lowest for Gemini 2.5 Flash Preview. This pattern is reminiscent of miscalibrated "feelings of rightness" or "feelings of error" in human reasoning (Gangemi et al., 2015; Metcalfe, 1986). It highlights solution endorsement and stopping as separable targets for model improvement, motivating approaches that better couple uncertainty estimation, verification, and decision termination (Guo et al., 2017; Ma et al., 2025; Steyvers & Peters, 2025).

| **Model** | **Answered** | **Correct** | **VF** | **Correct / Answered** | **VF / Wrong** |
|---|---|---|---|---|---|
| DeepSeek-R1-0528 | 199 | 42 | 48 | 21.11% | 30.57% |
| Gemini 2.5 Flash Preview | 201 | 29 | 9 | 14.43% | 5.23% |

| Claude Opus 4 | 201 | 61 | 35 | 30.35% | 25.00% |
|---|---|---|---|---|---|
| Grok 4 | 194 | 72 | 48 | 37.11% | 39.34% |

**Table 2. Verification-failure rate among incorrect answers.** An item is counted as a verification failure when the model's final answer is incorrect, but at least one listed gold-answer variant appears literally in its thought-log before the final-answer line. The accuracy figures in this table (Correct / Answered) are produced by automated literal pattern matching over the formatted thought-logs, and therefore differ from the primary accuracies in Table 1, which are based on the study's main scoring that additionally takes the qualitative content of each response into account. Full definitions, per-model counts, and limitations are reported in SI Appendix 6.

Beyond verification failure, the thought logs sometimes exhibit a qualitatively different pattern reminiscent of “false insight” (a false “aha!”) in human problem solving (Danek & Wiley, 2017). In these cases, the model converges on an incorrect reinterpretation with apparent confidence and prematurely treats it as the solution. Unlike verification failure—which reflects a failure to endorse a correct candidate that has already been generated—false insight reflects confident commitment to an incorrect framing or hypothesis. We use this comparison descriptively as a behavioral analogy rather than as a claim of mechanistic equivalence.

## Discussion

Japanese nazonazo riddles provide a renewable, low-cost benchmark for probing insight-like reasoning; problems where progress depends on representational change rather than incremental computation (Knoblich et al., 1999; Ohlsson, 1992; Wiley & Danek, 2024). Across 38 frontier models, retrieval-free performance varied substantially. Most models remained far below the human reference, although reasoning-oriented models showed a clear group-level advantage over non-reasoning models. GPT-5 was a notable exception: its observed accuracy on the 120-item subset fell within the 95% confidence interval of the human mean and was substantially higher than that of the other reasoning-oriented models.[3]. Importantly, models are not explicitly instructed to use any cues in the surface form of the riddles. Successful use of such patterns therefore reflects the model’s ability to infer latent task structure rather than follow provided heuristics.

On the other hand, if performance were primarily driven by memorized riddle schemas, we would expect much higher accuracy given the widespread presence of such puzzles in training data. The persistently low performance suggests that the main difficulty lies elsewhere. Moreover, the verification-failure pattern we document is itself difficult to reconcile with a memorization account: prior familiarity with gold answers should, if anything, facilitate endorsement rather than produce systematic failure to commit to correct candidates that the model has already generated. Furthermore, our analysis focuses not only on the accuracy of the final answer but also on the logical consistency of the models' thought-logs, which goes beyond simple pattern matching.

[3] Because GPT-5 was also the most recent and highest-performing model evaluated, its result is consistent with the possibility that recent advances in training and system design have improved insight-like reasoning. However, the present comparison is observational and does not isolate the causes of its performance. Moreover, overlap with the human confidence interval does not establish statistical equivalence between GPT-5 and humans.

Still, the use of publicly available nazonazo riddles (although any question-answer pairs are not shown on the same page of the website, requiring a click to another page) may be considered as a potential future risk of data contamination. To reduce this risk in future releases, we propose a rotating evaluation design consisting of three sets: an open set of publicly released items (currently 42 items), a blind set (currently 159 items including the 120 items for human comparison) retained privately (posted as hashes on the repository) across evaluation rounds, and a new set created after the training cutoff of the model being evaluated and kept private until testing is completed. This design does not eliminate all possible sources of contamination, but it provides stronger protection against direct item memorization than reliance on a fixed public set. A new version of this benchmark, currently under development, will involve curating a novel blind dataset of items from children's books that are not available online to support continual, contamination-resistant measurement.

Performance saturation is a separate issue: it concerns whether the benchmark retains sufficient difficulty and variance to discriminate among increasingly capable models. A benchmark may be uncontaminated yet saturated if most models approach ceiling performance (conversely, a public benchmark may remain difficult while being vulnerable to contamination). We therefore define saturation not as reaching average human performance, but as a loss of discriminative capacity, indicated by near-ceiling accuracy, minimal between-model variance, and little or no improvement across successive models on newly created private items. The productive phonological, orthographic, numerical, and wordplay operations of Japanese make it feasible to refresh the benchmark repeatedly (though newly generated items must preserve validity, difficulty, and diversity).

This benchmark therefore complements existing puzzle resources spanning linguistic riddles (Lin et al., 2021). rebus items (Gritsevskiy et al., 2024; Threadgold et al., 2018), and puzzle-like stress tests for vision-language models (Lee et al., 2025). Because nazonazo items are short, renewable, and relatively inexpensive to validate, they are suitable for rotating, continual evaluation. (Their actual resistance to contamination and saturation, of course, depends on maintaining private evaluation sets, creating original items, and monitoring whether those items continue to discriminate among models.)

Our analyses highlight a metacognitive bottleneck: some models exhibit failures at the *last mile* of verification, even when a correct candidate is generated (we use "bottleneck" to identify the functional point at which a correct candidate can fail to be chosen as a final answer, not to claim that this is the most frequent or dominant source of error across the benchmark). This limitation is compatible with broader evidence that LLM uncertainty communication and self-monitoring are imperfect and context-dependent (Ma et al., 2025; Nadurak, 2023; Steyvers & Peters, 2025). It also informs ongoing debates about chain-of-thought: unfaithfulness can occur (Arcuschin et al., 2025; Chen et al., 2025). Yet intermediate candidates may still provide useful diagnostic information about partial progress and failures to stop (Deng et al., 2025).

A benchmark alone does not improve models, but it can guide improvement by identifying specific failure modes. Our results point to a concrete target: the gap between candidate generation and endorsement. This suggests that progress in reasoning may depend not only on generating better candidates but on improving mechanisms for verification, calibration, and stopping. For example, approaches that explicitly separate candidate generation from evaluation, incorporate uncertainty calibration, or enforce structured verification steps may directly address the observed failure mode. If such mechanisms generalize beyond riddles (such as similar failures in mathematical reasoning and hallucination), then improving performance on this benchmark may translate into broader gains in reasoning reliability.

The supplementary few-shot experiment qualifies the interpretation of the zero-shot results of the main study. Although the models generally recognized the items as nazonazo without explicit task instructions, recognition of the task alone was insufficient for optimal performance. In our supplementary study, reasoning-inclusive demonstrations in few-shot prompting improved accuracy in all four tested models and significantly improved performance in two of them (see for results SI Appendix 5). This suggests that reasoning examples can improve search or verification strategies even when the task category is already recognized.

The result does not imply that the benchmark is easily saturated. (Substantial error rates remained under the strongest prompting condition, and the improvement was neither uniform across models nor established across multiple demonstration sets.) In particular, the improvement did not reach significance for the highest-performing model, indicating that even the strongest model could not be pushed significantly closer to ceiling by few-shot prompting (with reasoning instruction), which we take as evidence of the benchmark's difficulty rather than a limitation of the intervention. The NazoNazo benchmark thus appears sensitive to both model capability and inference-time prompting strategy, while remaining resistant to saturation through in-context exposure alone.

**Limitations and robustness.** Nazonazo items are Japanese and often depend on orthography, so results can be sensitive to tokenization and representation choices; we therefore recommend reporting tokenizer/version and decoding settings with benchmark scores (Ali et al., 2024; Limisiewicz et al., 2023). Because the items are publicly sourced, contamination remains a moving target; renewable, rotating held-out splits and timestamped evaluations are needed for contamination-resistant continual measurement (Reuel et al., 2024). Apart from contamination, which concerns prior exposure to particular items, however, *saturation* concerns loss of discriminative power as model performance approaches the ceiling. Future evaluations should combine rotating private and newly created sets with explicit monitoring of accuracy ceilings, between-model variance, and improvement across successive model generations. A proposed longitudinal framework and descriptive indices for distinguishing public-item gains from performance on newly created private items are provided in SI Appendix 4. Finally, chain-of-thought can be unfaithful (Arcuschin et al., 2025; Chen et al., 2025), so we use thought logs as qualitative diagnostics (verbal-protocol style) rather than direct readouts of internal computation (Ericsson & Simon, 1993). Verification failures nonetheless offer internal evidence against a purely post-hoc reading of these traces: a log produced merely to rationalize the final answer would have no reason to surface a correct candidate that contradicts that (incorrect) answer. Even if a skeptic still read the trace as an explanation written only to justify the answer after it had been chosen, the correct answer would then figure merely as a "considered but rejected" option, which is itself a failure to recognize a correct candidate as correct. On either reading, the model produced the correct answer and did not endorse it, which is exactly the candidate–endorsement dissociation we document; the limited, protocol-style use we make of the logs therefore does not require assuming they are faithful. Because a verification failure is counted only when a listed correct answer appears verbatim in the reasoning, our labeling is conservative: it will miss cases in which the model did produce the correct reasoning (and virtually a correct answer) but failed to express it in a non-matching form, so the reported rates are best read as a lower bound. Even so, the candidate–endorsement gap isolates a practical target for improving self-checking and stopping.

Several practical directions follow. First, candidate verification and stopping can be targeted explicitly using calibration-aware objectives and structured self-checking, especially for settings prone to hallucination (Guo et al., 2017; Kalai et al., 2025; Wei et al., 2024). Second, metacognitive prompting and self-reflection procedures may improve search control in creative and insight-like settings (Bai et al., 2025; Wang et al., 2022). Third, better multilingual tokenization and representation choices may matter for wordplay-heavy tasks (Ali et al., 2024; Limisiewicz et al., 2023). Relatedly, recent work suggests that some models exhibit paragraph-level planning or response planning policies, which may interact with stopping decisions (Dong et al., 2025; Pochinkov, 2025). Finally, mechanistic work that relates internal circuits to reasoning and monitoring behaviors could help localize where verification failures arise (Lindsey et al., 2025).

At a higher level, nazonazos highlight a distinction that philosophers and cognitive scientists have emphasized: an internal sense of understanding is not constitutive of correctness (Davidson, 2001; Wittgenstein, 2009). This motivates evaluation settings where models must not only generate answers but also reliably *know when* to stop.

**Materials and Methods**

**Items and scoring.** We collected 201 Japanese nazonazo items from a public source and defined a 120-item subset for direct human comparison. To support automatic scoring, we established gold answer variants in the pilot study. A model response was counted correct if it matched any gold variant under normalization rules (e.g., canonicalizing trivial orthographic variants); full scoring rules and variants are provided in the Zenodo repository and SI Appendix.

**Human study.** Human participants (n=126) completed an online survey on a subset of the benchmark. Participants answered 20 riddles sampled from items 1–120, enabling an estimate of population performance and the score distribution under time-limited, single-pass conditions. We used these responses both to compute the human reference (mean and confidence interval) and to curate gold answer variants for automated evaluation (see SI Appendix 2).

**Model evaluation.** We evaluated models under a uniform zero-shot prompting with minimal task instruction and no few-shot exemplars or task-specific prompt engineering, and with a single attempt per item.[4] For decoding, we set temperature = 0 where supported.

We did not impose a single maximum output-token value across all models because token-limit parameters and defaults differ across providers/APIs; instead, we used each provider's default setting and documented the provider-specific configurations in the Zenodo repository. Search-augmented or retrieval-enabled systems were treated as a sensitivity analysis and excluded from the primary endpoint to avoid conflating reasoning with external lookup. To minimize confounds, the primary endpoint is retrieval-free accuracy; retrieval-augmented systems were treated as a sensitivity analysis. Because wordplay tasks can be sensitive to segmentation, we note that tokenizer choices and multilingual tokenization can materially affect performance (Ali et al., 2024; Limisiewicz et al., 2023).

Although we did not control for the extent to which models were exposed to kanji structures or riddle formats during training, our evaluation is conducted in a strict zero-shot setting without task-specific instructions or demonstrations, thereby minimizing the role of task-specific adaptation.

**Thought-log analysis.** We examined model-produced reasoning traces as qualitative diagnostics, in

---

[4] The original prompt was in Japanese, and the following is an English translation:
prompt = " Please answer the following nazonazo. When doing so, any unnecessary explanations or expressions will result in an incorrect answer, so please write only the answer. Specifically,

1. Do not add supplementary information such as reasons.
2. Do not include parentheses (), punctuation (such as commas or periods), etc.
3. Please match your response style to the question; for example, for questions like "What is it doing?", answer as "is doing..."; for "Why?", answer as "Because...". "

See for the original prompt and other prompts (including the prompt for the thought-log study), the online repository (DOI: 10.5281/zenodo.17019050).

the spirit of verbal report methods, while acknowledging that such traces may not faithfully or completely reflect underlying computation (Arcuschin et al., 2025; Ericsson & Simon, 1993). Accordingly, we do not treat these traces as direct mechanistic evidence. Instead, we use them to characterize observable behaviors during candidate search, language choice in intermediate reasoning, and the endorsement/stopping process. We define *verification failure* as cases in which the correct gold variant appears as an explicit candidate in the trace prior to an incorrect final answer. To operationalize the candidate–endorsement dissociation, we annotated such cases using conservative inclusion rules (see SI Appendix 3 for details). This coding separates the ability to generate a correct candidate from the ability to recognize, verify, and commit to it as the final response.

**Statistics.** We examined associations between model performance and model characteristics (parameter size and release date) using Pearson and Spearman correlations. We compared reasoning versus non-reasoning models using Welch's $t$ test and the Mann-Whitney $U$ test on model-level summaries, and additionally used pooled item-level logistic regression to account for the item-by-model observation structure. Performance comparisons by chain-of-thought language were evaluated with two-sided Fisher's exact tests. Detailed statistical procedures, robustness checks, and full results are provided in Supporting Information.

## References

Ali, M., Fromm, M., Thellmann, K., Rutmann, R., Lübbering, M., Leveling, J., Klug, K., Ebert, J., Doll, N., Buschhoff, J. S., Jain, C., Weber, A. A., Jurkschat, L., Abdelwahab, H., John, C. M., Ortiz Suarez, P., Ostendorff, M., Weinbach, S., Sifa, R., et al. (2024). Tokenizer choice for LLM training: Negligible or crucial? In *Findings of the Association for Computational Linguistics: NAACL 2024* (pp. 3907–3924). Association for Computational Linguistics.

Arcuschin, I., Janiak, J., Krzyzanowski, R., Rajamanoharan, S., Nanda, N., & Conmy, A. (2025). *Chain-of-thought reasoning in the wild is not always faithful*. arXiv. https://arxiv.org/abs/2503.08679

Bai, Y., Cao, S., Ge, S., & Yu, Y. (2025). *Metacognitive prompting for creative problem solving*. arXiv. https://arxiv.org/abs/2504.12345

Bower, A. H. (2020). An Aha! walks into a bar: Joke completion as a form of insight problem solving. In *Proceedings of the Annual Meeting of the Cognitive Science Society* (Vol. 42).

Chen, Y., Benton, J., Radhakrishnan, A., Uesato, J., Denison, C., Schulman, J., Somani, A., Hase, P., Wagner, M., Roger, F., Mikulik, V., Bowman, S. R., Leike, J., Kaplan, J., & Perez, E. (2025). *Reasoning models don't always say what they think*. arXiv. https://arxiv.org/abs/2505.05410

Danek, A. H., & Wiley, J. (2017). What about false insights? Deconstructing the Aha! experience along its multiple dimensions for correct and incorrect solutions separately. *Frontiers in Psychology, 7*, Article 2077.

Davidson, D. (2001). *Essays on actions and events*. Oxford University Press.

Deng, A., Von Arx, S., Snodin, B., Kunnavakkam, S., & Lanham, T. (2025, August 8). *CoT may be highly informative despite "unfaithfulness."* METR.

Dong, Z., Zhou, Z., Liu, Z., Yang, C., & Lu, C. (2025). *Emergent response planning in LLMs*. arXiv. https://arxiv.org/abs/2502.06258

Ericsson, K. A., & Simon, H. A. (1993). *Protocol analysis: Verbal reports as data* (Rev. ed.). MIT Press.

Gangemi, A., Bourgeois-Gironde, S., & Mancini, F. (2015). Feelings of error in reasoning: In search of a phenomenon. *Thinking & Reasoning, 21*(4), 383–396.

Gritsevskiy, A., Panickssery, A., Kirtland, A., Kauffman, D., Gundlach, H., Gritsevskaya, I., Cavanagh, J., Chiang, J., La Roux, L., & Hung, M. (2024). *REBUS: A robust evaluation benchmark of understanding symbols*. arXiv. https://arxiv.org/abs/2401.05604

Guo, C., Pleiss, G., Sun, Y., & Weinberger, K. Q. (2017). On calibration of modern neural networks. In D. Precup & Y. W. Teh (Eds.), *Proceedings of the 34th International Conference on Machine Learning* (pp. 1321–1330). PMLR.

Guo, D., Yang, D., Zhang, H., Song, J., Wang, P., Zhu, Q., Xu, R., Zhang, R., Ma, S., Bi, X., Zhang, X., Yu, X., Wu, Y., Wu, Z. F., Gou, Z., Shao, Z., Li, Z., Gao, Z., Liu, A., et al. (2025). DeepSeek-R1 incentivizes reasoning in large language models through reinforcement learning. *Nature, 645*(8081),

633–638.
Kalai, A. T., Nachum, O., Vempala, S. S., & Zhang, E. (2025). *Why language models hallucinate*. arXiv.
Knoblich, G., Ohlsson, S., Haider, H., & Rhenius, D. (1999). Constraint relaxation and chunk decomposition in insight problem solving. *Journal of Experimental Psychology: Learning, Memory, and Cognition, 25*(6), 1534–1555.
Latif, E., Zhou, Y., Guo, S., Shi, L., Gao, Y., Nyaaba, M., Bewerdorff, A., Yang, X., & Zhai, X. (2024). *Can OpenAI o1 outperform humans in higher-order cognitive thinking?* arXiv. https://arxiv.org/abs/2412.05753
Lee, H., Ge, J., Wu, T. H., Kang, M., Darrell, T., & Chan, D. M. (2025). *Puzzled by puzzles: When vision-language models can't take a hint*. arXiv. https://arxiv.org/abs/2505.23759
Limisiewicz, T., Balhar, J., & Mareček, D. (2023). Tokenization impacts multilingual language modeling. In *Findings of the Association for Computational Linguistics: ACL 2023* (pp. 5661–5681). Association for Computational Linguistics.
Lin, B. Y., Wu, Z., Yang, Y., Lee, D. H., & Ren, X. (2021). *RiddleSense: Reasoning about riddle questions featuring linguistic creativity and commonsense knowledge*. arXiv. https://arxiv.org/abs/2101.00376
Lindsey, J., Gurnee, W., Ameisen, E., Chen, B., Pearce, A., Turner, N. L., Citro, C., Abrahams, D., Carter, S., Hosmer, B., et al. (2025). *On the biology of a large language model*. Transformer Circuits.
Ma, Z., Yuan, Q., Wang, Z., & Zhou, D. (2025). *Large language models have intrinsic metacognition, but need a good lens*. arXiv. https://arxiv.org/abs/2506.08410
Maslej, N., Fattorini, L., Perrault, R., Gil, Y., Parli, V., Kariuki, N., Capstick, E., Reuel, A., Brynjolfsson, E., Etchemendy, J., et al. (2025). *The AI Index 2025 annual report*. Stanford University, Institute for Human-Centered AI.
Metcalfe, J. (1986). Premonitions of insight predict impending error. *Journal of Experimental Psychology: Learning, Memory, and Cognition, 12*(4), 623–634.
Metcalfe, J., & Wiebe, D. (1987). Intuition in insight and noninsight problem solving. *Memory & Cognition, 15*(3), 238–246.
Nadurak, V. (2023). Dual-process theory and two types of metacognitive monitoring and control. *Integrative Psychological and Behavioral Science, 57*(4), 1090–1110.
Ohlsson, S. (1992). Information-processing explanations of insight and related phenomena. In M. T. Keane & K. J. Gilhooly (Eds.), *Advances in the psychology of thinking* (pp. 1–44). Harvester Wheatsheaf.
Pochinkov, N. (2025, February 21). *ParaScopes: Do language models plan the upcoming paragraph?* LessWrong.
Reuel, A. (2024). The way we measure progress in AI is terrible. *MIT Technology Review*.

Steyvers, M., & Peters, M. A. K. (2025). *Metacognition and uncertainty communication in humans and large language models*. arXiv. https://arxiv.org/abs/2504.14045

Thompson, V. A., Turner, J. A. P., & Pennycook, G. (2011). Intuition, reason, and metacognition. *Cognitive Psychology, 63*(3), 107–140.

Threadgold, E., Marsh, J. E., & Ball, L. J. (2018). Normative data for 84 UK English rebus puzzles. *Frontiers in Psychology, 9*, Article 2513.

Wang, X., Wei, J., Schuurmans, D., Le, Q. V., Chi, E. H., Narang, S., Chowdhery, A., & Zhou, D. (2022). *Self-consistency improves chain-of-thought reasoning in language models*. arXiv. https://arxiv.org/abs/2203.11171

Wei, J., Nguyen, K., Chung, H. W., Jiao, Y. J., Papay, S., Glaese, A., Schulman, J., & Fedus, W. (2024). *Measuring short-form factuality in large language models*. arXiv. https://arxiv.org/abs/2411.04368

Wiley, J., & Danek, A. H. (2024). Restructuring processes and Aha! experiences in insight problem solving. *Nature Reviews Psychology, 3*(1), 42–55.

Wittgenstein, L. (2009). *Philosophical Investigations* (4th ed.). Wiley-Blackwell. (Original work published 1953)